\documentclass[acmtog, authorversion, nonacm]{acmart}
\pdfoutput=1
\AtBeginDocument{%
  \providecommand\BibTeX{{%
    \normalfont B\kern-0.5em{\scshape i\kern-0.25em b}\kern-0.8em\TeX}}}

\acmJournal{FACMP}

\usepackage[percent]{overpic}

\newcommand{\ourmethod}{MotionCLIP}

\newcommand{\tsl}{@{\hskip 4.\tabcolsep}}
\newcommand{\etal}{et al.}

\newcommand\blfootnote[1]{%
  \begingroup
  \renewcommand\thefootnote{}\footnote{#1}%
  \addtocounter{footnote}{-1}%
  \endgroup
}

\newif\ifeccv
\eccvfalse

\ifeccv
\newcommand{\scite}[1]{\cite{#1}}
\else
\newcommand{\scite}[1]{\shortcite{#1}}
\fi

\setcitestyle{nosort,square}
\begin{document}
\title{\ourmethod{}: Exposing Human Motion Generation to CLIP Space}

\author{Guy Tevet$^*$}\thanks{\;\;\; * The authors contributed equally}
\affiliation{\institution{Tel Aviv University} \country{Israel}}
\author{Brian Gordon$^*$}
\affiliation{\institution{Tel Aviv University} \country{Israel}} 
\author{Amir Hertz}
\affiliation{\institution{Tel Aviv University} \country{Israel}}
\author{Amit H. Bermano}
\affiliation{\institution{Tel Aviv University} \country{Israel}}
\author{Daniel Cohen-Or}
\affiliation{\institution{Tel Aviv University} \country{Israel}}

\ifeccv
\begin{figure}
\centering
\begin{overpic}[width=1\columnwidth,tics=10, trim=0mm 0 0mm 0,clip]{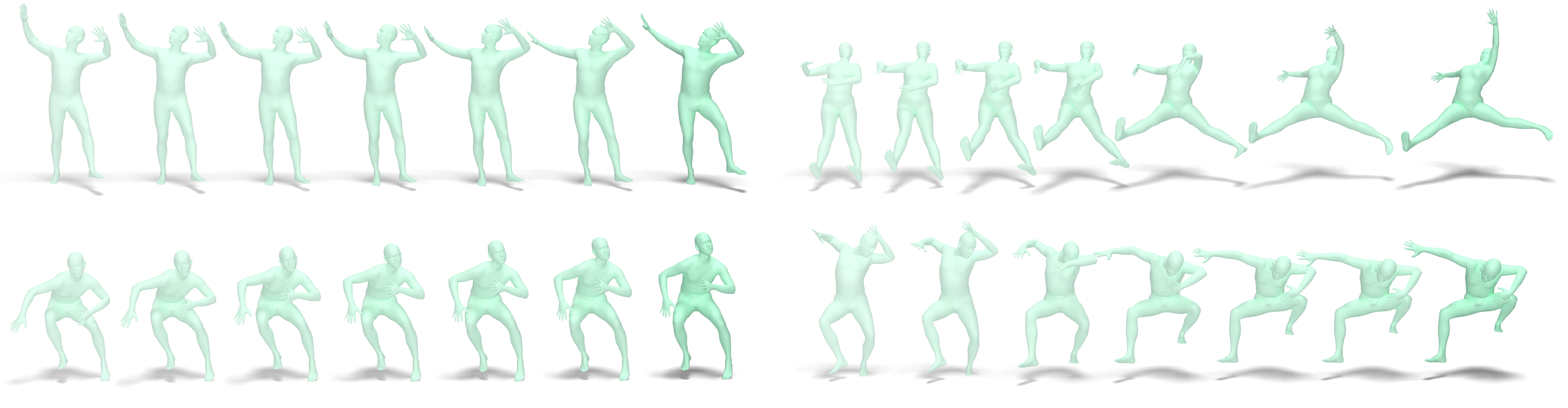}
     \put(20,11.5){``Usain Bolt"}
     \put(23,-1){``Gollum"}
     \put(68,11.5){``Swan lake"}
     \put(63,-1){``Spiderman in action!"}
\end{overpic}
\caption{Motions generated by \ourmethod{} conditioned on 
different cultural references.
\ourmethod{} exploits the rich knowledge encapsulated in pre-trained language-images model (CLIP) and projects the human motion manifold over its latent space.
}
\label{fig:teaser}
\end{figure}
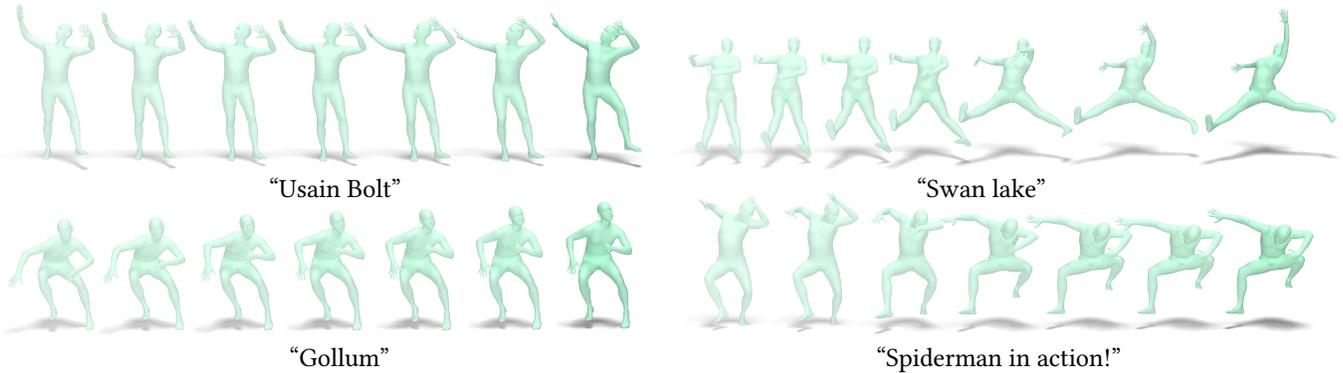
\else
\begin{teaserfigure}
\centering
\large
\begin{overpic}[width=1\columnwidth,tics=10, trim=0mm 0 0mm 0,clip]{figures/teaser-05.png}
     \put(20,11.5){``Usain Bolt''}
     \put(21.5,-1){``Gollum''}
     \put(68,11.5){``Swan lake''}
     \put(65,-1){``Spiderman in action!''}
\end{overpic}
\caption{Motions generated by \ourmethod{} conditioned on 
different cultural references.
\ourmethod{} exploits the rich knowledge encapsulated in pre-trained language-images model (CLIP) and projects the human motion manifold over its latent space.
}
\label{fig:teaser}
\end{teaserfigure}
 \fi

\begin{abstract}
We introduce \ourmethod{}, a 3D human motion auto-encoder featuring a latent embedding that is disentangled, well behaved, and supports highly semantic textual descriptions.
\ourmethod{} gains its unique power by aligning its latent space with that of the Contrastive Language-Image Pre-training (CLIP) model.  Aligning the human motion manifold to CLIP space implicitly infuses the extremely rich semantic knowledge of CLIP into the manifold. In particular, it helps continuity by placing semantically similar motions close to one another, and disentanglement, which is inherited from the CLIP-space structure.
\ourmethod{} comprises a transformer-based motion auto-encoder, trained to reconstruct motion while being aligned to its text label's position in CLIP-space. 
We further leverage CLIP's unique visual understanding and inject an even stronger signal through aligning motion to rendered frames in a self-supervised manner.
We show that although CLIP has never seen the motion domain, \ourmethod{} offers unprecedented text-to-motion abilities, allowing out-of-domain actions, disentangled editing, and abstract language specification. For example, the text prompt ``couch" is decoded into a sitting down motion, due to lingual similarity, and the prompt ``Spiderman" results in a web-swinging-like solution that is far from seen during training.  
In addition, we show how the introduced latent space can be leveraged for motion interpolation, editing and recognition.
\footnote{See our project page: \url{https://guytevet.github.io/motionclip-page/}}
\blfootnote{* The authors contributed equally}

\end{abstract}


\maketitle

\section{Introduction}

Human motion generation includes the intuitive description, editing, and generation of 3D sequences of human poses. It is relevant to many applications that require virtual or robotic characters.
Motion generation is, however, a challenging task.
Perhaps the most challenging aspect is the limited availability of data, which is expensive to acquire and to label.
Recent years have brought larger sets of motion capture acquisitions~\cite{AMASS:ICCV:2019}, sometimes sorted by classes~\cite{liu2019ntu,ji2018large} or even labeled with free text~\cite{BABEL:CVPR:2021,plappert2016kit}.
Yet, it seems that while this data may span a significant part of human motion, it is not enough for machine learning algorithms to understand the semantics of the motion manifold, and it is definitely not descriptive enough for natural language usage. 
Hence, neural models trained using labeled motion data \cite{ahuja2019language2pose,lin2018generating,yamada2018paired,petrovich21actor,maheshwari2022mugl} do not generalize well to the full richness of the human motion manifold, nor to the natural language describing it.

\ifeccv
\begin{figure*}[t!]
\centering
\scriptsize
\begin{overpic}[width=.9\textwidth,tics=10, trim=0mm 0 0mm 0,clip]{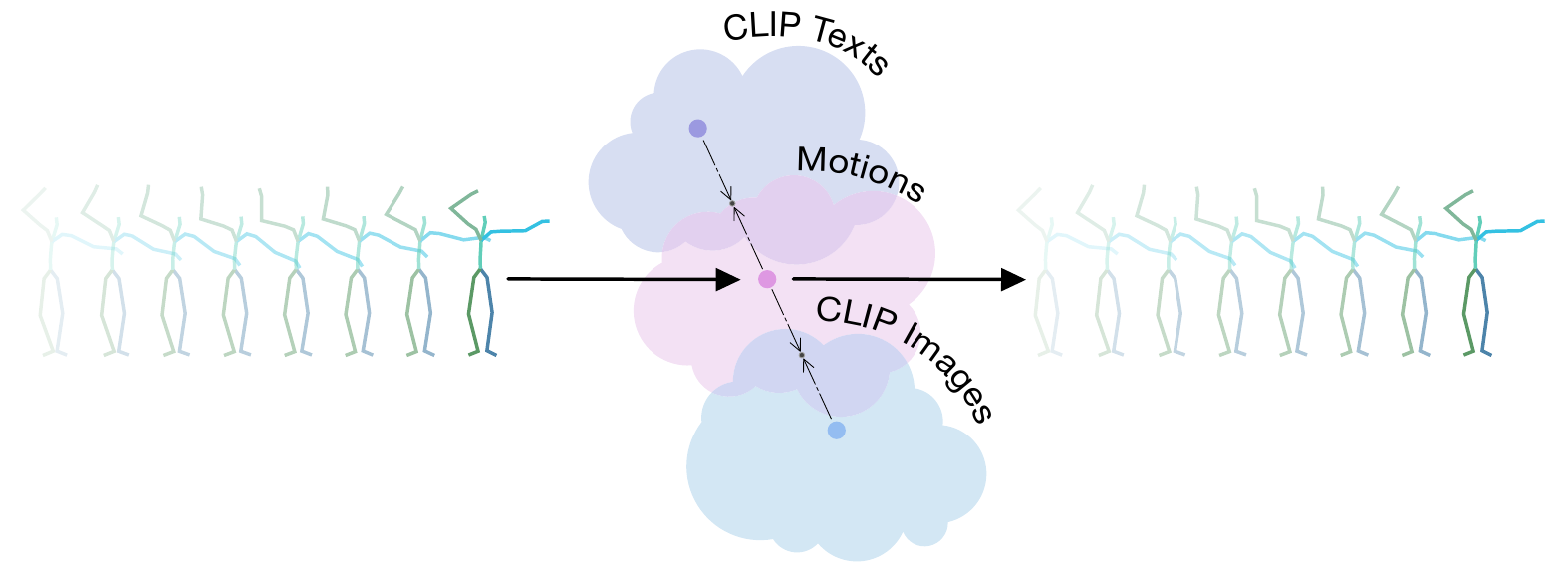}
     \put(12,11.2){Motion}
     \put(68,11.2){Reconstructed Motion}
     
     \put(40,23.5){$\mathcal{L}_\textnormal{text}$}
     \put(43,14){$\mathcal{L}_\textnormal{image}$}
\end{overpic}
\caption{\ourmethod{} overview. A motion auto-encoder is trained to simultaneously reconstruct motion sequences while aligning their latent representation with corresponding texts and images representations in CLIP space.}
\label{fig:overview}
\end{figure*}
\else
\begin{figure*}[t!]
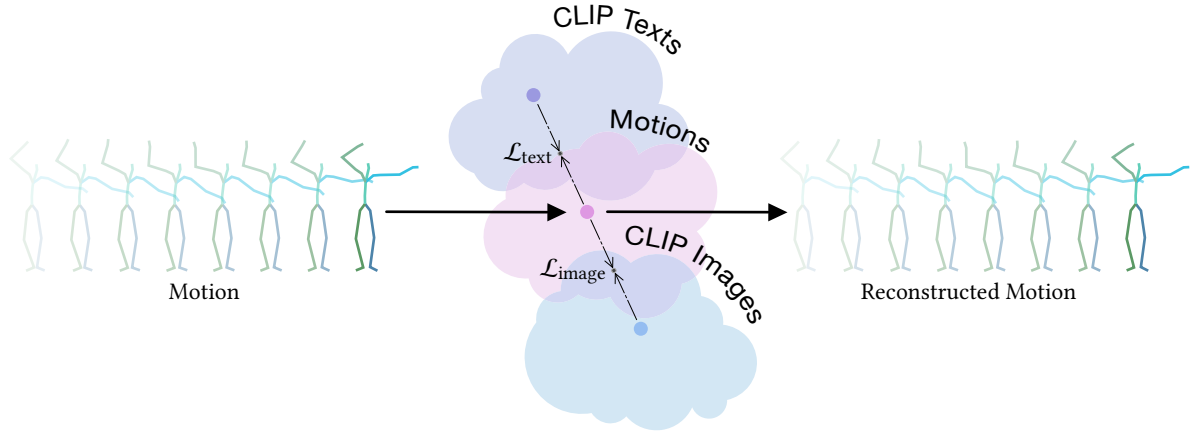

\centering
\begin{overpic}[width=.9\textwidth,tics=10, trim=0mm 0 0mm 0,clip]{figures/diagram_c.pdf}
     \put(14.5,12){Motion}
     \put(71.5,12){Reconstructed Motion}
     
     \put(42,23.5){$\mathcal{L}_\textnormal{text}$}
     \put(45,14){$\mathcal{L}_\textnormal{image}$}
\end{overpic}
\caption{\ourmethod{} overview. A motion auto-encoder is trained to simultaneously reconstruct motion sequences while aligning their latent representation with corresponding texts and images representations in CLIP space.}
\label{fig:overview}
\end{figure*}
\fi

In this work, we introduce \ourmethod{}, a 3D motion auto-encoder that induces a latent embedding that is disentangled, well behaved, and supports highly semantic and elaborate descriptions. To this end, we employ CLIP~\cite{radford2021learning}, a large scale visual-textual embedding model.
Our key insight is that even though CLIP has not been trained on the motion domain what-so-ever, we can inherit much of its latent space's %
virtue by enforcing its powerful and semantic structure onto the motion domain. To do this, we train a transformer-based~\cite{vaswani2017attention} auto-encoder that is aligned to the latent space of CLIP, using existing motion textual labels. In other words, we train an encoder to find the proper embedding of an input sequence in CLIP space, and a decoder that generates the most fitting motion to a given CLIP space latent code. 
To further improve the alignment with CLIP-space, we also leverage CLIP's visual encoder, and synthetically render frames to guide the alignment in a self-supervised manner (see Figure~\ref{fig:overview}). As we demonstrate, this step is crucial for out-of-domain generalization, since it allows finer-grained description of the motion, unattainable using text.

The merit of aligning the human motion manifold to CLIP space is two-fold: 
First, combining the geometric motion domain with lingual semantics benefits the semantic description of motion. As we show, this benefits tasks such as text-to-motion and motion style transfer. More importantly however, we show that this alignment benefits the motion latent space itself, infusing it with semantic knowledge and inherited disentanglement.
Indeed, our latent space demonstrates unprecedented compositionality of independent actions, semantic interpolation between actions, 
and even natural and linear latent-space based editing.

As mentioned above, the textual and visual CLIP encoders offer the semantic description of motion. In this aspect, our model demonstrates never-before-seen capabilities for the field of motion generation. For example, motion can be specified using arbitrary natural language, through abstract scene or intent descriptions instead of the motion directly, or even through pop-culture references. For example, the CLIP embedding for the phrase 
``wings" is decoded into a flapping motion like a bird,
and ``Williams sisters"  into a tennis serve, 
since these terms are encoded close to motion seen during training, thanks to CLIP's semantic understanding. Through the compositionality induced by the latent space, the aforementioned process also yields clearly unseen motions, such as the iconic web-swinging gesture that is produced for the input "Spiderman" (see this and other culture references in Figure~\ref{fig:teaser}). Our model also naturally extents to other downstream tasks. In this aspect, we depict motion interpolation to depict latent smoothness, editing to demonstrate disentanglement, and action recognition to point out the semantic structure of our latent space.
For all these applications, we show comparable or preferable results either through metrics or a user study, even though each task is compared against a method that was designed especially for it. Using the action recognition benchmark, we also justify our design choices with an ablation study.

\section{Related Work}

\subsection{Guided Human Motion Generation}

One means to guide motion generation is to condition on another domain.
An immediate, but limited, choice is conditioning on \emph{action} classes.  ACTOR~\cite{petrovich21actor} and Action2Motion~\cite{guo2020action2motion} suggested learning this multi-modal distribution from existing action recognition datasets using Conditional Variational-Autoencoder(CVAE) \cite{sohn2015learning} architectures. MUGL~\cite{maheshwari2022mugl} model followed with elaborated Conditional Gaussian-Mixture-VAE~\cite{dilokthanakul2016deep} that supports up to $120$ classes and multi-person generation, based on the NTU-RGBD-120 dataset~\cite{liu2019ntu}.

Motion can be conditioned on other domains. %
For example, recent works~\cite{li2021dance,aristidou2021rhythm} generated dance moves conditioned on music and the motion prefix. Edwards~\etal~\scite{edwards2016jali} generated facial expressions to fit a speaking audio sequence.

A more straightforward approach to control motion is using another motion. In particular, for style transfer applications. Holden~\etal~\scite{holden2016deep} suggested to code style  using the latent code's Gram matrix, inspired by Gatys~\etal~\scite{Gatys_2016_CVPR}. Aberman~\etal~\scite{aberman2020unpaired} injected style attributes using a dedicated temporal-invariant AdaIN layer~\cite{huang2017arbitrary}. Recently, Wen~\etal~\scite{wen2021autoregressive} encoded style in the latent code of Normalizing Flow generative model \cite{dinh2014nice}.
We show that \ourmethod{} also encodes style in its latent representation, without making any preliminary assumptions or using a dedicated architecture.

\begin{figure*}[t!]
\centering
\begin{overpic}[width=.9\textwidth,tics=10, trim=0mm 0 0mm 0,clip]{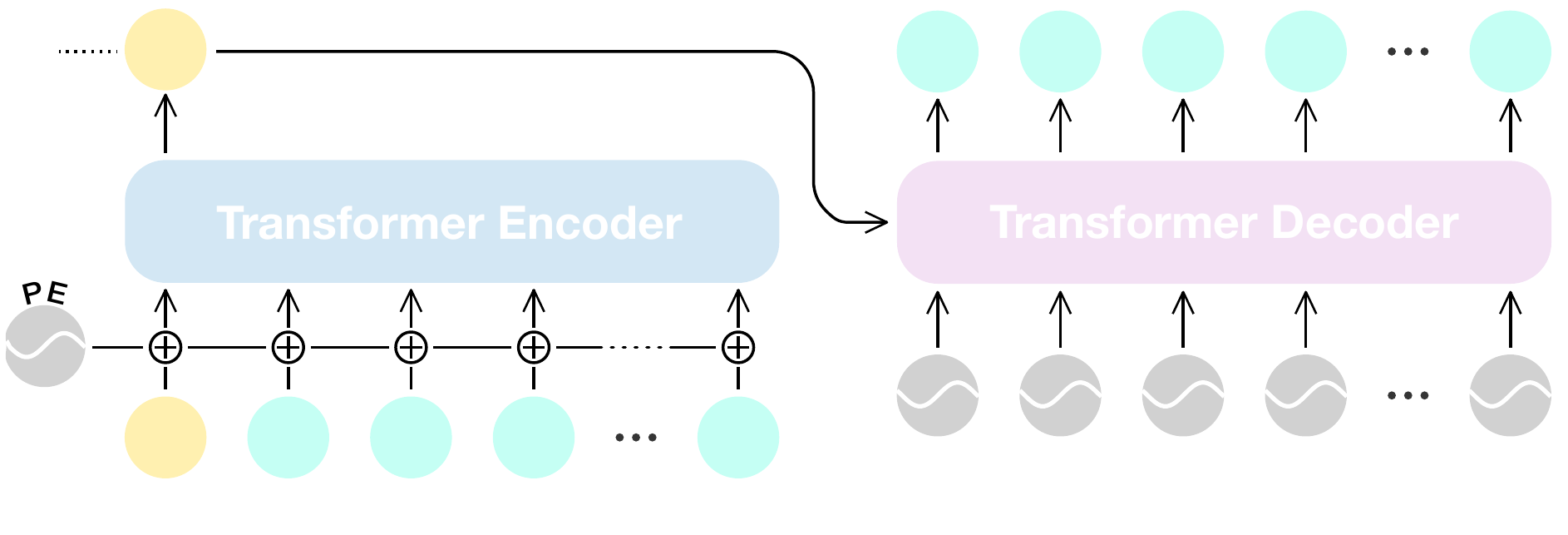}
     \put(9.3,5.9){$z_{tk}$}
     \put(9.7,30.6){$z_{p}$}
     \put(17.6, 5.9){$p_{1}$}
     \put(25.3,5.9){$p_{2}$}
     \put(33.2,5.9){$p_{3}$}
     \put(46,5.9){$p_{T}$}

     \put(59, 30.4){$\hat{p}_{1}$}
     \put(66.8,30.4){$\hat{p}_{2}$}
     \put(74.7,30.4){$\hat{p}_{3}$}
     \put(82.5,30.4){$\hat{p}_{4}$}
     \put(95.4,30.4){$\hat{p}_{T}$}
     
     \put(57.6, 4.1){$t=1$}
     \put(67.2,4.1){$2$}
     \put(75.1,4.1){$3$}
     \put(82.9,4.1){$4$}
     \put(95.6,4.1){$T$}
     
     \put(-.5,32.5){$\mathcal{L}_\textnormal{text}$}
     \put(1.5,30.3){$+$}
     \put(-.8,28.5){$\mathcal{L}_\textnormal{image}$}

\end{overpic}
\caption{Motion Auto-Encoder. A transformer encoder is trained to project a motion sequence $p_{1:T}$ into a latent vector $z_p$ in CLIP latent space. Simultaneously, a transformer decoder is trained to recover the motion by attending to $z_p$. 
}
\label{fig:architecture}
\end{figure*}

\subsection{Text-to-Motion}
The KIT dataset\cite{plappert2016kit} provides about 11 hours of motion capture sequences, each sequence paired with a sentence explicitly describing the action performed. KIT sentences describe the action type, direction and sometimes speed, but lacks details about the style of the motion, and not including abstract descriptions of motion. Current text-to-motion research is heavily based on KIT. Plappert~\etal~\scite{plappert2018learning} learned text-to-motion and motion-to-text using seq2seq RNN-based architecture. 
Yamada~\etal~\scite{yamada2018paired} learned those two mappings by simultaneously training text and motion auto-encoders while binding their latent spaces using text and motion pairs. Lin~\etal~\scite{lin2018generating} further improved trajectory prediction by adding a dedicated layer. Ahuja~\etal~\scite{ahuja2019language2pose} introduced JL2P model, which got improved results with respect to nuanced concepts of the text, namely velocity, trajectory and action type. They learned joint motion-text latent space and apply training curriculum to ease optimization.

More recently, BABEL dataset~\cite{BABEL:CVPR:2021} provided per-frame textual labels ordered in $260$ classes to the larger AMASS dataset~\cite{AMASS:ICCV:2019}, including about 40 hours of motion capture. Although providing explicit description of the action, often lacking any details besides the action type, this data spans a larger variety of human motion. 
\ourmethod{}  overcomes the data limitations by leveraging out-of-domain knowledge using CLIP~\cite{radford2021learning}.

\subsection{CLIP aided Methods} 

Neural networks have successfully learned powerful latent representations coupling natural images with natural language describing it~\cite{he2017fine,ramesh2021zero}.
A recent example is CLIP\cite{radford2021learning}, a model coupling images and text in deep latent space using a constructive objective\cite{hadsell2006dimensionality,chen2020simple}. By training over hundred millions of images and their captions, CLIP gained a reach semantic latent representation for visual content.
This expressive representation enables high quality image generation and editing, controlled by natural language~\cite{patashnik2021styleclip,gal2021stylegan,frans2021clipdraw}.
Even more so, this model has shown that connecting the visual and textual worlds also benefits purely visual tasks~\cite{vinker2022clipasso}, simply by providing a well-behaved, semantically structured, latent space. %

Closer to our method are works that utilize the richness of CLIP outside the imagery domain. In the 3D domain, CLIP's latent space provides a useful objective that enables semantic manipulation \cite{sanghi2021clip,text2mesh,wang2021clip} where the domain gap is closed by a neural rendering.
CLIP is even adopted in temporal domains \cite{guzhov2021audioclip,Luo2021CLIP4Clip,fang2021clip2video} that utilize large datasets of video sequences that are paired with text and audio.
Unlike these works that focus on classification and retrieval, we introduce a generative approach that utilizes limited amount of human motion sequences that are paired with text.

\section{Method}
\label{sec:method}

Our goal is learning a semantic and disentangled motion representation that will serve as a basis for generation and editing tasks. To this end, we need to learn not only the mapping to this representation (encoding), but also the mapping back to explicit motion (decoding). 

Our training process is illustrated in Figure~\ref{fig:overview}.
We train a transformer-based motion auto-encoder, while aligning the latent motion manifold to CLIP joint representation.
We do so using (i) a \textit{Text Loss}, connecting motion representations to the CLIP embedding of their text labels, and (ii) an \textit{Image Loss}, connecting motion representations to CLIP embedding of rendered images that depict the motion visually.

At inference time, semantic editing applications can be performed in latent space. For example,
to perform style transfer, we find a latent vector representing the style, and simply add it to the content motion representation and decode the result back into motion. Similarly, to classify an action, we can simply encode it into the latent space, and see to which of the class text embedding it is closest.
Furthermore, we use the CLIP text encoder to perform text-to-motion - An input text is decoded using the text encoder then directly decoded by our motion decoder. 
The implementation of these and other applications is detailed in Section~\ref{sec:results}.

We represent motion sequences using the SMPL body model~\cite{loper2015smpl}. A sequence of length $T$ denoted $p_{1:T}$ such that 
$p_i \in \mathbb{R}^{24 \times 6}$ defines orientations in 6D representation\cite{zhou2019continuity} for global body orientation and 23 SMPL joints, at the $i^\textnormal{th}$ frame. 
The mesh vertices locations $v_{1:T}$ are calculated according to SMPL specifications with $\beta=0$ and a neutral-gender body model following Petrovich et al.~\cite{petrovich21actor}.

\begin{figure*}[ht]
\centering
\includegraphics[width=\textwidth]{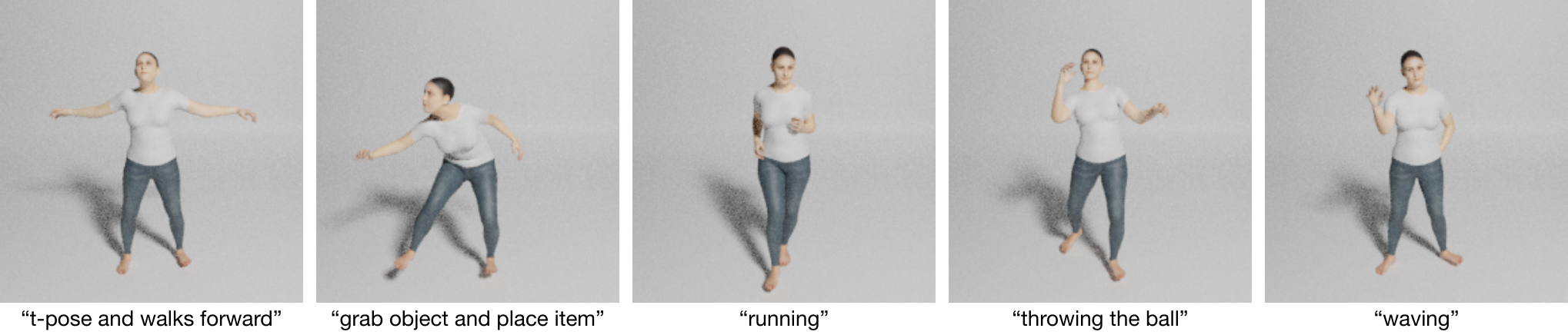}
\caption{A sample of the rendered frames and their text description used during training.}
\label{fig:render}
\end{figure*}

To project the motion manifold into the latent space, we learn a transformer-based auto-encoder \cite{vaswani2017attention}, adapted to the motion domain \cite{petrovich21actor,wang2021multi,li2021dance}. \ourmethod{}'s architecture is detailed in Figure~\ref{fig:architecture}.

\textbf{Transformer Encoder.} $E$, Maps a motion sequence $p_{1:T}$ to its latent representation $z_p$. The sequence is embedded into the encoder's dimension by applying linear projection for each frame separately, then adding standard positional embedding. The embedded sequence is the input to the transformer encoder, together with additional learned prefix token $z_{tk}$. The latent representation, $z_p$ is the first output (the rest of the sequence is dropped out). Explicitly, $z_p = E(z_{tk}, p_{1:T})$.

\textbf{Transformer Decoder.} $D$, predicts a motion sequence $\hat{p}_{1:T}$ given a latent representation $z_p$. This representation is fed to the transformer as key and value, while the query sequence is simply the positional encoding of $1:T$. The transformer outputs a representation for each frame, which is then mapped to pose space using a linear projection. Explicitly, $\hat{p}_{1:T} = D(z_p)$.
We further use a differentiable SMPL layer to get the mesh vertices locations, $\hat{v}_{1:T}$.

\textbf{Losses.} This auto-encoder is trained to represent motion via reconstruction $L2$ losses on 
joint orientations, joint velocities and vertices locations. 
Explicitly,

\begin{equation} \label{eq1}
\begin{split}
    \mathcal{L}_\textnormal{recon} = \frac{1}{|p|T} \sum_{i =1}^{T} \| p_i - \hat{p}_i\|^{2} + 
    \frac{1}{|v|T} \sum_{i =1}^{T} \| v_i - \hat{v}_i\|^{2} \\ + \frac{1}{|p|(T-1)} \sum_{i =1}^{T-1} \| (p_{i+1} - p_i) - (\hat{p}_{i+1} - \hat{p}_{i})\|^{2}
\end{split}
\end{equation}

Given text-motion and image-motion pairs, $(p_{1:T}, t)$, $(p_{1:T}, s)$ correspondingly, we attach the motion representation to the text and image representations using cosine distance,

\begin{equation}
    \mathcal{L}_\textnormal{text} = 1 - \cos(CLIP_\textnormal{text}(t), z_p)
\end{equation}
and
\begin{equation}
    \mathcal{L}_\textnormal{image} = 1 - \cos(CLIP_\textnormal{image}(s), z_p)
\end{equation}

The motion-text pairs can be derived from labeled motion dataset, whereas the images can be achieved
by rendering a single pose from a motion sequence, to a synthetic image $s$, in an unsupervised manner (More details in Section~\ref{sec:results}).

Overall, the loss objective of \ourmethod{} is defined,

\begin{equation}
    \mathcal{L} = \mathcal{L}_\textnormal{recon} + \lambda_\textnormal{text} \mathcal{L}_\textnormal{text} + \lambda_\textnormal{image} \mathcal{L}_\textnormal{image}
\end{equation}

\section{Results}
\label{sec:results}

To evaluate \ourmethod{}, we consider its two main advantages. In Section~\ref{sec:text2motion}, we inspect \ourmethod{}'s ability to convert text into motion. Since the motion's latent space is aligned to that of CLIP, we use CLIP's pretrained text encoder to process input text, and convert the resulting latent embedding into motion using \ourmethod{}'s decoder. We compare our results to the state-of-the-art and report clear preference for both seen and unseen generation. We also show comparable performance to state-of-the-art style transfer work simply by adding the style as a word to the text prompt. Lastly, we exploit CLIP expert lingual understanding to convert 
abstract text into corresponding, and sometimes unexpected, motion. 

In Section~\ref{sec:manifold_applications} we focus on the resulting auto-encoder, and the properties of its latent-space. We inspect its smoothness and disentanglement. Smoothness is shown through well-behaved interpolations, even between distant motion. Disentanglement is demonstrated using latent space arithmetic; by adding and subtracting various motion embeddings, we achieve compositionality and semantic editing. Lastly, we leverage our latent structure to perform action recognition over the trained encoder. The latter setting is also used for ablation study. In the following, we first lay out the data used, and other general settings.

\subsection{General Settings}

We train our model on the BABEL dataset~\cite{BABEL:CVPR:2021}. It comprises about 40 hours of motion capture data, represented with the SMPL body model~\cite{loper2015smpl}. Each frame is annotated with per-frame textual labels, and is categorized into one of 260 action classes. We down sample the data to $30$ frames per-second and cut it into sequences of length $60$. We get a single textual label per sequence by listing all actions in a given sequence, then concatenating them to a single string. Finally, we choose for each motion sequence a random frame to be rendered using the \emph{Blender} software and the SMPL-X add-on~\cite{SMPL-X:2019} (See Figure~\ref{fig:render}). This process outputs triplets of (motion, text, synthetic image) which are used for training.

We train a transformer auto-encoder with $8$ layers for each encoder and decoder as described in Section~\ref{sec:method}. We align it with the \emph{CLIP-ViT-B/32} frozen model. Out of the data triplets, the text-motion pairs are used for the \emph{text loss} and image-motion pairs for the \emph{image loss}. Both $\lambda$ values are set to $0.01$ throughout our experiments.
\footnote{\url{https://github.com/GuyTevet/MotionCLIP}}

\begin{figure*}
\centering
\includegraphics[width=.95\textwidth]{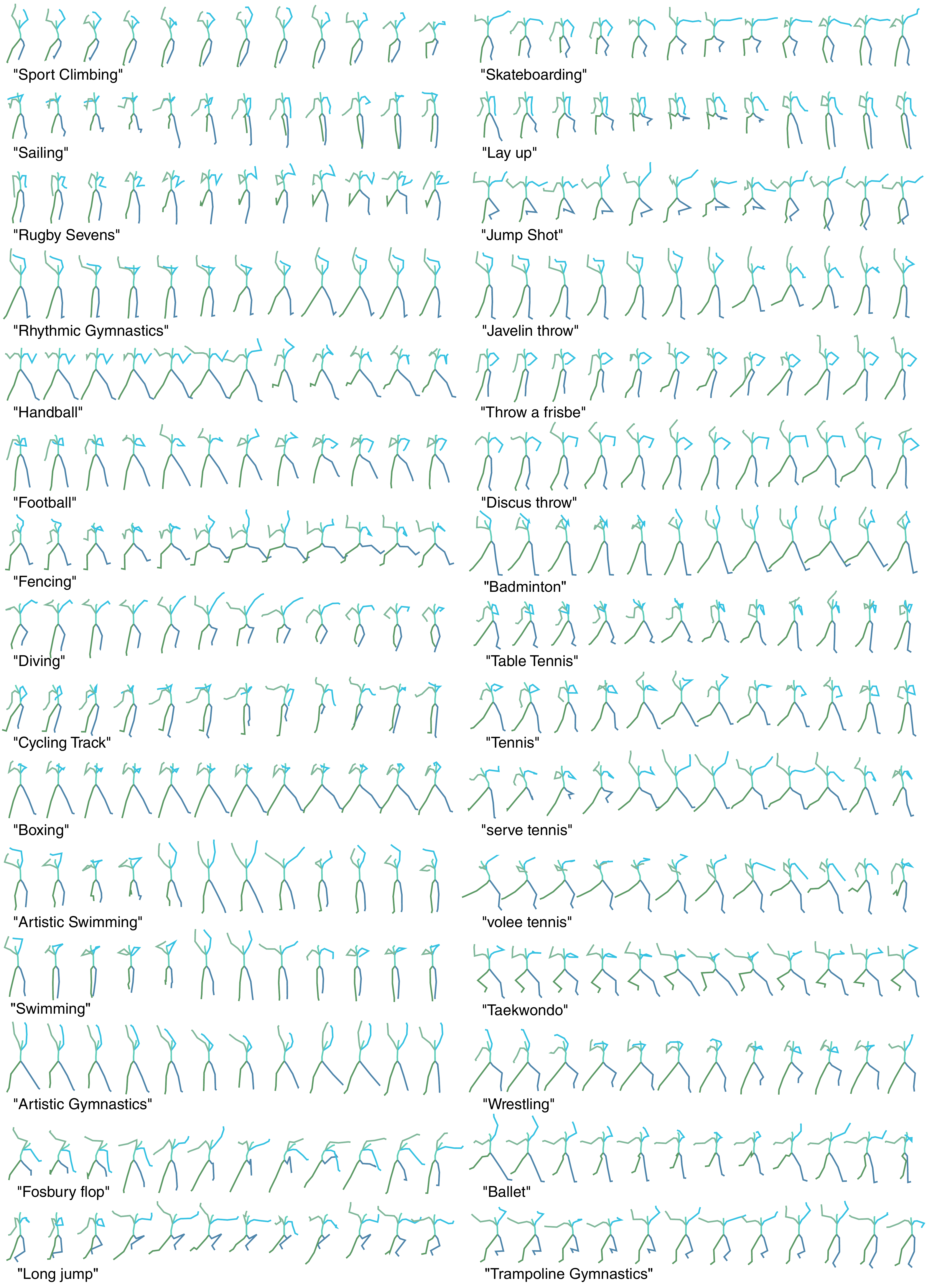}
\caption{Sport motions generated by \ourmethod{} conditioned on the text beneath each row. }
\label{fig:sports}
\end{figure*}

\begin{figure*}[t!]
\centering
\begin{overpic}[width=\textwidth,tics=10, trim=0mm 0 0mm 0,clip]{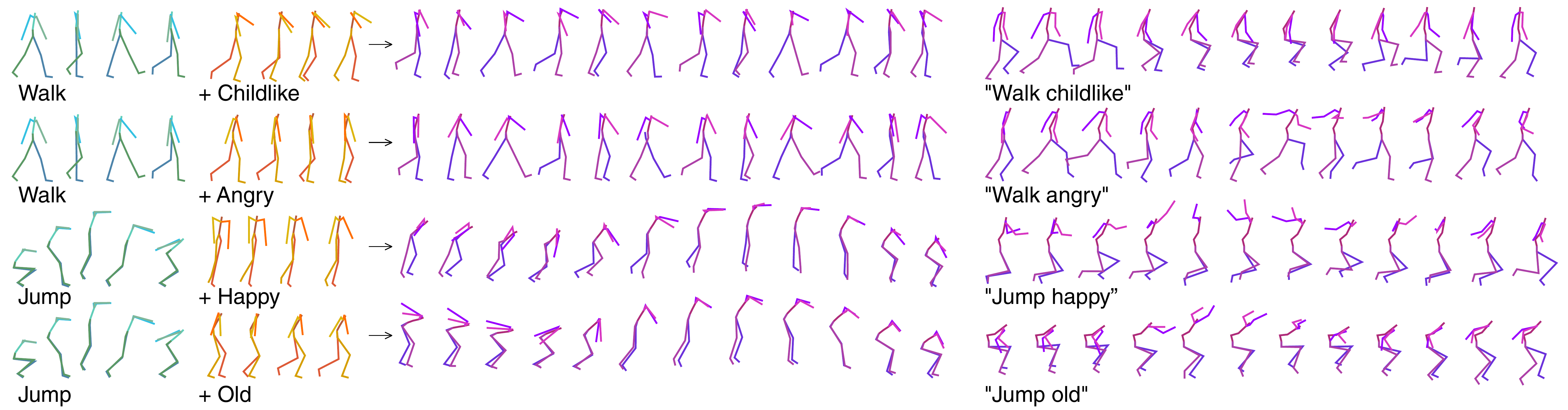}
\put(25,-1){Aberman et al.~\shortcite{aberman2020unpaired}}
\put(76,-1){\ourmethod{}}
\end{overpic}
\vspace{1pt}
\caption{Style generation. Left: style transfer by Aberman~\etal~\shortcite{aberman2020unpaired}, conditioned on action (green) and style (orange) motions. Right: \ourmethod{} generating style from plain text input.}
\label{fig:style_user_study}
\end{figure*}

\subsection{Text-to-Motion}
\label{sec:text2motion}

\emph{Text-to-motion} is performed at inference time, using the CLIP text encoder and \ourmethod{} decoder, without any further training.
Even though not directly trained for this task, \ourmethod{} shows unprecedented performance in text-to-motion, dealing with explicit descriptions, subtle nuances and abstract language.

\textbf{Actions.}
We start by demonstrating the capabilities of \ourmethod{} to generate explicit actions - both seen and unseen in training. We compare our model to JL2P~\cite{ahuja2019language2pose}. Since the two models were trained on different datasets, we define a new common ground for evaluation. 
We define two new sets of samples for a user study:
(1)The \emph{in-domain set} comprises actions with textual labels that appear in at least $0.5\%$ of the labels of both datasets, and (2) the \emph{Out-of-domain set} includes textual labels that do not appear in any of the labels of both datasets, hence, unseen for both models. For fairness, we construct this set from the list of Olympic sports (both summer and winter) that are disjoint to both datasets. 
We conduct a user study, comparing the generation of each model conditioned on a given textual label. For each example, we then ask users to choose which of the two motions best fits the label. 
Table~\ref{table:actions} shows that \ourmethod{} was clearly preferred by the users for both sets. Figure~\ref{fig:sports} demonstrates a variety of sports performed by \ourmethod{}, as used in the user-study. Note how even though this is not a curated list, the motion created according to all 30 depicted text prompts resembles the requested actions.

\begin{figure*}
\centering
\includegraphics[width=.98\textwidth]{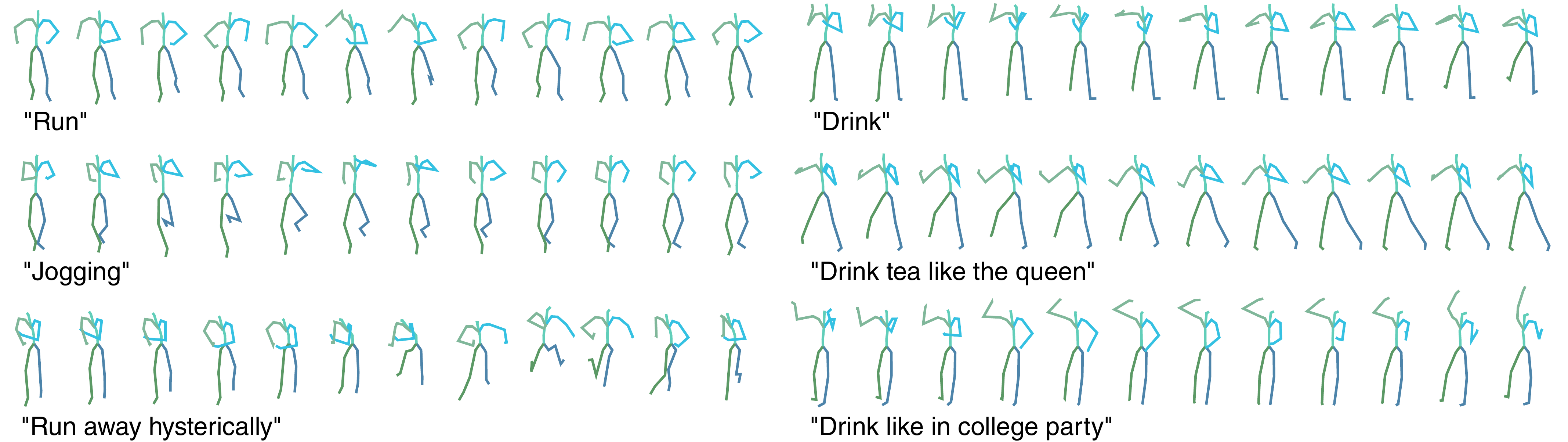}
\caption{\ourmethod{} expresses the style described as a free text.}
\label{fig:free_style}
\end{figure*}

\textbf{Styles.}
We investigate \ourmethod{}'s ability to represent motion style, without being explicitly trained for it. 
We compare the results produced by \ourmethod{} to the style transfer model by Aberman~\etal~\scite{aberman2020unpaired}. The latter receives two input motion sequences, one indicating content and the other style, and combines them through a dedicated architecture, explicitly trained to disentangle style and content from a single sequence. In contrast, we simply feed \ourmethod{} with the action and style textual names (e.g.``walk proud"). We show to users the outputs of the two models side-by-side and ask them to choose which one presents both style and/or action better (See Figure~\ref{fig:style_user_study}). Even though Aberman~\etal{} was trained specifically for this task and gets the actual motions as an input, rather then text, Table~\ref{table:style} shows comparable results for the two models, with an expected favor toward Aberman~\etal{}. This, of course, also means that \ourmethod{} allows expressing style with free text, and does not require an exemplar motion to describe it. Such novel free text style augmentations are demonstrated in Figure~\ref{fig:free_style}.

\begin{figure*}[]
\centering
\includegraphics[width=.98\textwidth]{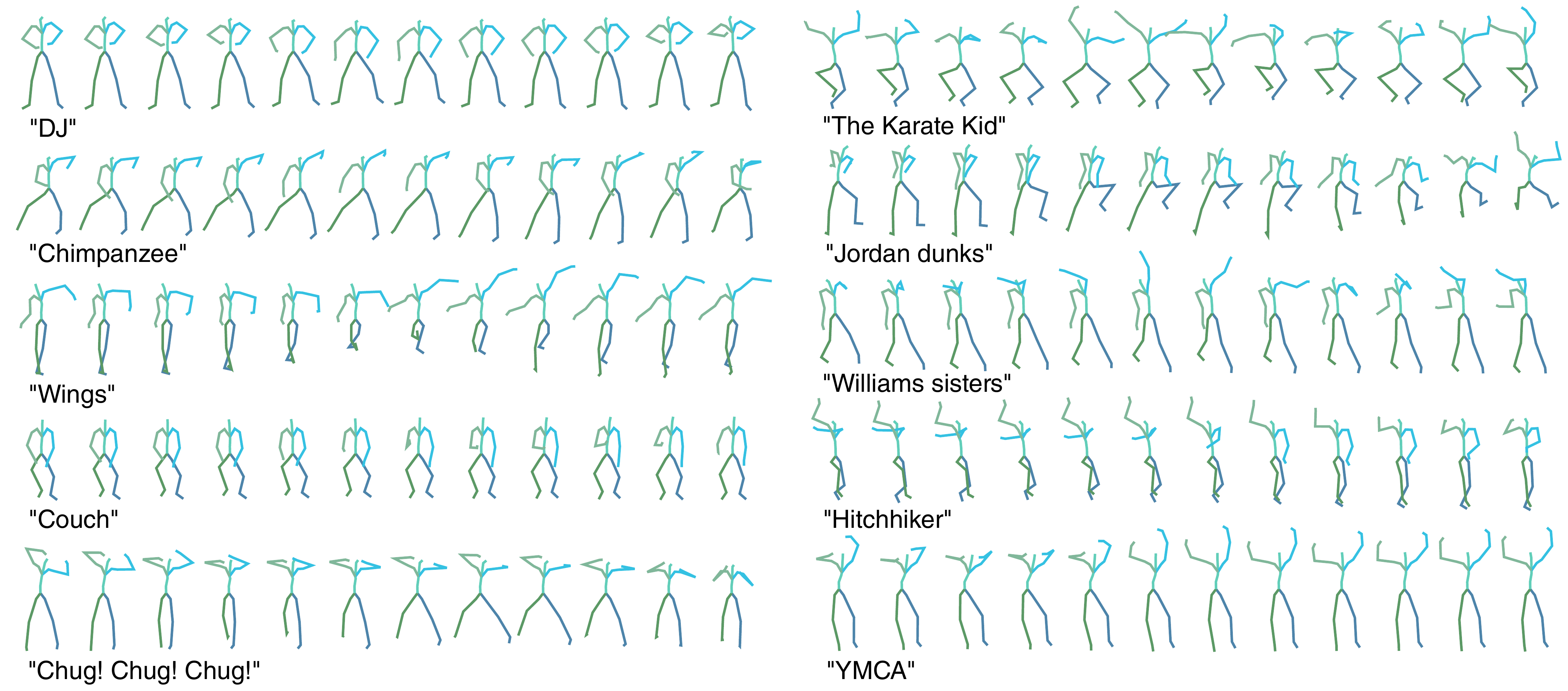}
\caption{Abstract language. \ourmethod{} generates the signature motions of culture figures and phrases.}
\label{fig:abstract_language}
\end{figure*}

\begin{figure*}[h]
\centering
\includegraphics[width=\textwidth]{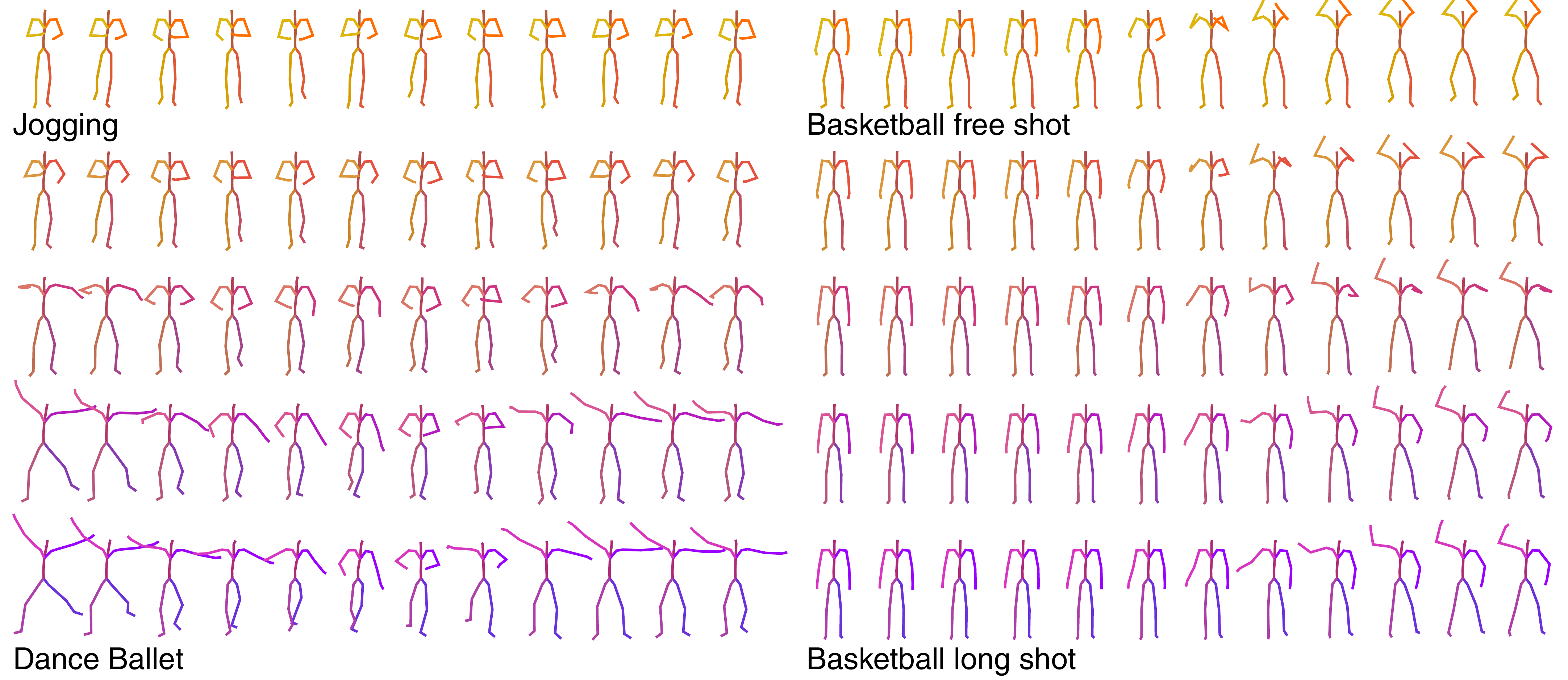}
\caption{Latent space motion interpolation. \ourmethod{} enables semantic interpolation between two motions.}
\label{fig:interp}
\end{figure*}

\begin{figure*}[h]
\centering
\includegraphics[width=\textwidth]{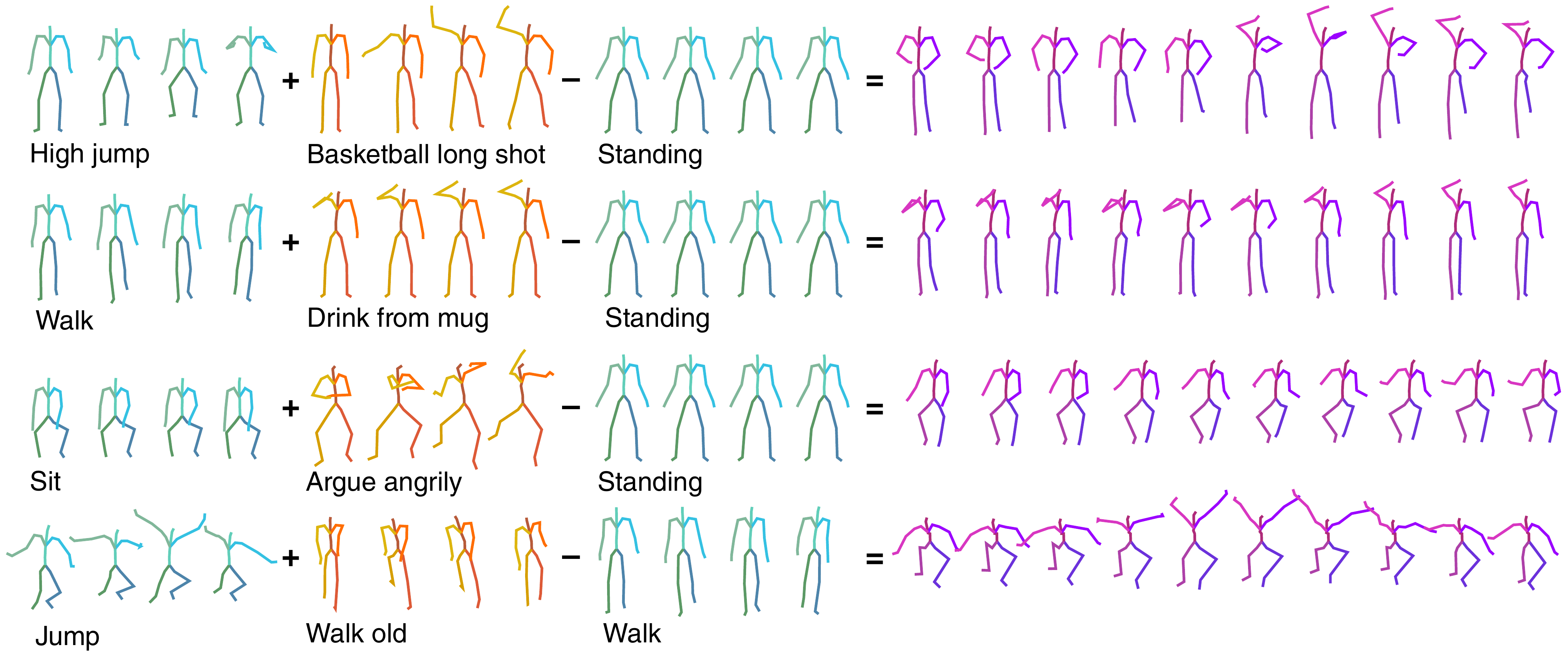}
\caption{Latent space motion editing. \ourmethod{} enables semantic editing in latent space. Here we demonstrate two applications (1) upper and lower body action compositions (top two examples) and (2) style transfer (the two examples at the bottom).}
\label{fig:edit}
\end{figure*}

\textbf{Abstract language.}
One of the most exciting capabilities of \ourmethod{} is generating motion given text that doesn't explicitly describe motion. This includes obvious linguistic connections, such as the act of sitting down, produced from the input text "couch". Other, more surprising examples include mimicking the signature moves of famous real and fictional figures, like \emph{Usain Bolt} and \emph{The Karate Kid}, and other cultural references like the famous ballet performance of \emph{Swan Lake} and the \emph{YMCA} dance (Figures~\ref{fig:teaser} and ~\ref{fig:abstract_language}). These results include motions definitely not seen during training (e.g., Spiderman in Figure~\ref{fig:teaser}), which strongly indicates how well the motion manifold is aligned to CLIP space.

\subsection{Motion Manifold Applications}
\label{sec:manifold_applications}

It is already well established that the CLIP space is smooth and expressive. We demonstrate its merits also exist in the aligned motion manifold, through the following experiments.

\textbf{Interpolation}
As can be seen in Figure~\ref{fig:interp}, linear interpolation between two latent codes yields semantic transitions between motions in both time and space. This is a strong indication to the smoothness of this representation. 
Here, the source and target motions (top and bottom respectively) were sampled from the validation set, and between them are three transitions evenly sampled from the linear trajectory between the two motion representations, then decoded by \ourmethod{}.

\textbf{Latent-Based Editing}
To demonstrate how disentangled and uniform \ourmethod{} latent space is, we experiment with latent-space arithmetic to edit motion (see Figure~\ref{fig:edit}). As can be seen, these linear operations allow motion compositionality - the upper body action can be decomposed from the lower body one, and recomposed with another lower body performance. In addition, Style can be added by simply adding the vector of the style name embedding. These two properties potentially enable intuitive and semantic editing even for novice users. 

\begin {table}[h]

\centering

\begin{tabular}{lcc \tsl  cc}
   \toprule
& \multicolumn{2}{c}{\textbf{JL2P~\shortcite{ahuja2019language2pose}}} & \multicolumn{2}{c}{\textbf{\ourmethod{}}}  \\
& seen  &  &  seen &  \\
& in train & pref &  in train & pref \\
 \hline
  In-domain set & yes & 23.3\% & yes & \textbf{76.7\%} \\
  Out-of-domain set & no & 24.7\% & no &  \textbf{75.3\%} \\

\bottomrule
\end{tabular}

\vspace{5pt}
\caption{Action generation from text - user study. \emph{pref} is the preference score of each model (when compared side-by-side). \emph{Seen in train} notes whether or not the samples are taken from a distribution seen by each model during train. 
\ourmethod{} is clearly preferred by the users.
}
\label{table:actions}
\end {table}

\begin {table}[h]

\centering

\begin{tabular}{l c \tsl \tsl c}
   \toprule
& \textbf{Aberman~\etal~\shortcite{aberman2020unpaired}}  & \textbf{\ourmethod{}} \\
 \hline
  Happy &    31.3\% & \textbf{68.7\%}  \\
  Proud &  \textbf{86.4\%} &  13.6\% \\
  Angry &  43.5\% &   \textbf{56.5\%} \\
  Childlike &  \textbf{57.6\%} &   42.4\% \\
  Depressed &  \textbf{74.2\%} &   25.8\% \\
  Drunk &  \textbf{50\%} &   \textbf{50\%} \\
  Old &  \textbf{57.7\%} &   42.3\% \\
  Heavy &  \textbf{85.2\%} &   14.8\% \\
 \hline
  Average &  \textbf{62.1\%} &   37.9\% \\ 
\bottomrule
\end{tabular} 

\vspace{5pt}
\caption{Style generation - user study (preference score side-by-side). We compare our style + action generation from text, to those of Aberman et al.~\scite{aberman2020unpaired} which gets style and content motions as input. Interestingly, although not 
trained to generate style, our model wins twice and break even once}
\label{table:style}
\end {table}

\begin {table}[ht]
\centering

\begin{tabular}{l  c  c} 
    \toprule
    & Top-1 acc. & Top-5 acc. \\
    \midrule
    \ourmethod{} & 40.9 \% & 57.71\%\\
    W.O. image loss & 35.05\% & 50.26\%\\
    W.O. text loss & 4.54\% & 18.37\%\\
    2s-AGCN \shortcite{shi2019two} & 41.14\% & 73.18\%\\
    
 \bottomrule
\end{tabular}

\vspace{5pt}
\caption{Action Recognition. Using \ourmethod{} together with CLIP text encoder for classification yields performance marginally close to 2s-AGCN~\cite{shi2019two} dedicated architecture on the BABEL-60 benchmark.}
\label{table:action_recognition}
\end {table}

\textbf{Action Recognition}
Finally, we further demonstrate how well our latent spaces is semantically structured. We show how combined with the CLIP text encoder, \ourmethod{} encoder can be used for action recognition. We follow BABEL $60$-classes benchmark and train the model with BABEL class names instead of the raw text. At inference, we measure the cosine distance of a given motion sequence to all $60$ class name encodings and apply softmax, as suggested originally for image classification~\cite{radford2021learning}. In table~\ref{table:action_recognition}, we compare Top-1 and Top-5 accuracy of \ourmethod{} classifier to 2s-AGCN classifier~\cite{shi2019two}, as reported by Punnakkal~\etal~\scite{BABEL:CVPR:2021}. As can be seen, this is another example where our framework performs similarly to dedicated state-of-the-art methods, even though \ourmethod{} was not designed for it.

\section{Conclusions}
\label{sec:conclusions}

We have presented a motion generation network that leverages the knowledge encapsulated in CLIP, allowing intuitive operations, such as text conditioned motion generation and editing. As demonstrated, training an auto-encoder on the available motion data alone struggles to generalize well, possibly due to data quality or the complexity of the domain. Non the less, we see that the same auto-encoder with the same data can lead to a significantly better understanding of the motion manifold and its semantics, merely by aligning it to a well-behaved knowledge-rich latent space. 

We restress the fascinating fact that even though CLIP has never seen anything from the motion domain, or any other temporal signal, its latent structure naturally induces semantics and disentanglement. This succeeds even though the connection between CLIP's latent space and the motion manifold is through sparse and inaccurate textual labeling. In essence, the alignment scheme transfers semantics by encouraging the encoder to place semantically similar samples closer together. Similarly, it induces the disentanglement built into the CLIP space, as can be seen, for example, in our latent-space arithmetic experiments.

Of course, \ourmethod{} has its limitations, opening several novel research opportunities.
It struggles to understand directions, (e.g. left, right and counter-clockwise), to capture some styles (such as heavy and proud), and is of course not consistent for out-of-domain cultural reference exapmles (e.g, it fails to produce \emph{Cristiano Ronaldo}'s goal celebration, and \emph{Superman}'s signature pose).
Nonetheless, we believe \ourmethod{} is an important step toward intuitive motion generation. Knowledge-rich disentangled latent spaces have already proven themselves as a flexible tool to novice users in other fields, such as facial images. In the future, we would like to further explore how powerful large-scale latent spaces could be leveraged to benefit additional domains. We would also like to explore more elaborate architectures and domain adaptation schemes for the main generation part, and to deepen our investigation into downstream tasks that could benefit from this powerful backbone.

\pagebreak

\bibliographystyle{ACM-Reference-Format}
\bibliography{motionclip}

\end{document}